\documentclass[lettersize,journal]{IEEEtran}
\usepackage{amsmath,amsfonts}
\usepackage{algorithmic}
\usepackage{algorithm}
\usepackage{array}
\usepackage[caption=false,font=normalsize,labelfont=sf,textfont=sf]{subfig}
\usepackage{textcomp}
\usepackage{stfloats}
\usepackage{url}
\usepackage{verbatim}
\usepackage{graphicx}
\usepackage{cite}
\usepackage{bbm}
\usepackage{multirow}
\usepackage{dsfont}
\usepackage{arydshln}
\usepackage{color}

\begin{document}

\title{LAPP: Layer Adaptive Progressive Pruning for Compressing CNNs from Scratch}

\author{Pucheng Zhai, Kailing Guo$^{\ast}$, ~\IEEEmembership{Member,~IEEE,} Fang Liu, ~\IEEEmembership{Member,~IEEE,} \\Xiaofen Xing, ~\IEEEmembership{Member,~IEEE,} Xiangmin Xu, ~\IEEEmembership{Senior Member,~IEEE,}
\thanks{Pucheng Zhai, and Xiaofen Xing are with South China University of Technology, Guangzhou 510640, China (email:pucheng\_zhai@163.com; xfxing@scut.edu.cn). Kailing Guo and Xiangmin Xu are with South China University of Technology, Guangzhou 510640, China, and with the Pazhou Lab, Guangzhou 510330, China (email:guokl@scut.edu.cn; xmxu@scut.edu.cn). Fang Liu is with Guangdong University of Finance, Guangzhou 510521, China (e-mail: 47-032@gduf.edu.cn). }
\thanks{$^{\ast}$Corresponding author.}}

\markboth{\tiny{This work has been submitted to the IEEE for possible publication. Copyright may be transferred without notice, after which this version may no longer be accessible.}}%
{Shell \MakeLowercase{\textit{et al.}}: A Sample Article Using IEEEtran.cls for IEEE Journals}


\maketitle

\begin{abstract}
Structured pruning is a commonly used convolutional neural network (CNN) compression approach. Pruning rate setting is a fundamental problem in structured pruning. Most existing works introduce too many additional learnable parameters to assign different pruning rates across different layers in CNN or cannot control the compression rate explicitly. Since too narrow network blocks information flow for training, automatic pruning rate setting cannot explore a high pruning rate for a specific layer. To overcome these limitations, we propose a novel framework named Layer Adaptive Progressive Pruning (LAPP), which gradually compresses the network during initial training of a few epochs from scratch. In particular, LAPP designs an effective and efficient pruning strategy that introduces a learnable threshold for each layer and FLOPs constraints for network. Guided by both task loss and FLOPs constraints, the learnable thresholds are dynamically and gradually updated to accommodate changes of importance scores during training. Therefore the pruning strategy can gradually prune the network and automatically determine the appropriate pruning rates for each layer. What's more, in order to maintain the expressive power of the pruned layer, before training starts, we introduce an additional lightweight bypass for each convolutional layer to be pruned, which only adds relatively few additional burdens. Our method demonstrates superior performance gains over previous compression methods on various datasets and backbone architectures. For example, on CIFAR-10, our method compresses ResNet-20 to 40.3$\%$ without accuracy drop. 55.6$\%$ of FLOPs of ResNet-18 are reduced with 0.21$\%$ top-1 accuracy increase and 0.40$\%$ top-5 accuracy increase on ImageNet. 
\end{abstract}


\section{Introduction}
\IEEEPARstart{I}{N} recent years, convolutional neural networks (CNNs) have made remarkable achievements in many computer vision applications such as classification \cite{simonyan2014very}, \cite{he2016deep}, object detection \cite{redmon2016you}, \cite{lin2017focal}, face recognition \cite{marriott20213d}, \cite{li2021spherical}, action recognition \cite{chen2021channel}, and semantic segmentation \cite{nirkin2021hyperseg}. However, while performance of CNN models continues to get better, the CNNs become deeper and wider, which results in an explosive growth in the parameters and FLOPs. Thus, existing CNNs require very high storage and computational costs, which makes it difficult for CNNs to be deployed on small devices with limited resources. In order to solve this problem, CNN compression techniques have attracted more and more attention. Common CNN compression techniques include pruning \cite{li2016pruning,ople2021adjustable,wang2020pruning}, low-rank tensor decomposition \cite{lebedev2015speeding,phan2020stable,kossaifi2020factorized}, knowledge distillation \cite{hinton2015distilling, park2021learning}, low-bit quantization \cite{sharma2021generalized,li2021residual,duan2022differential}, and compact architecture design \cite{han2022ghostnets, zhang2021split,chen2023run}. In this paper, we focus on pruning which is a research hotspot.

\begin{figure*}[!t]
\centering
\includegraphics[width=6.0in]{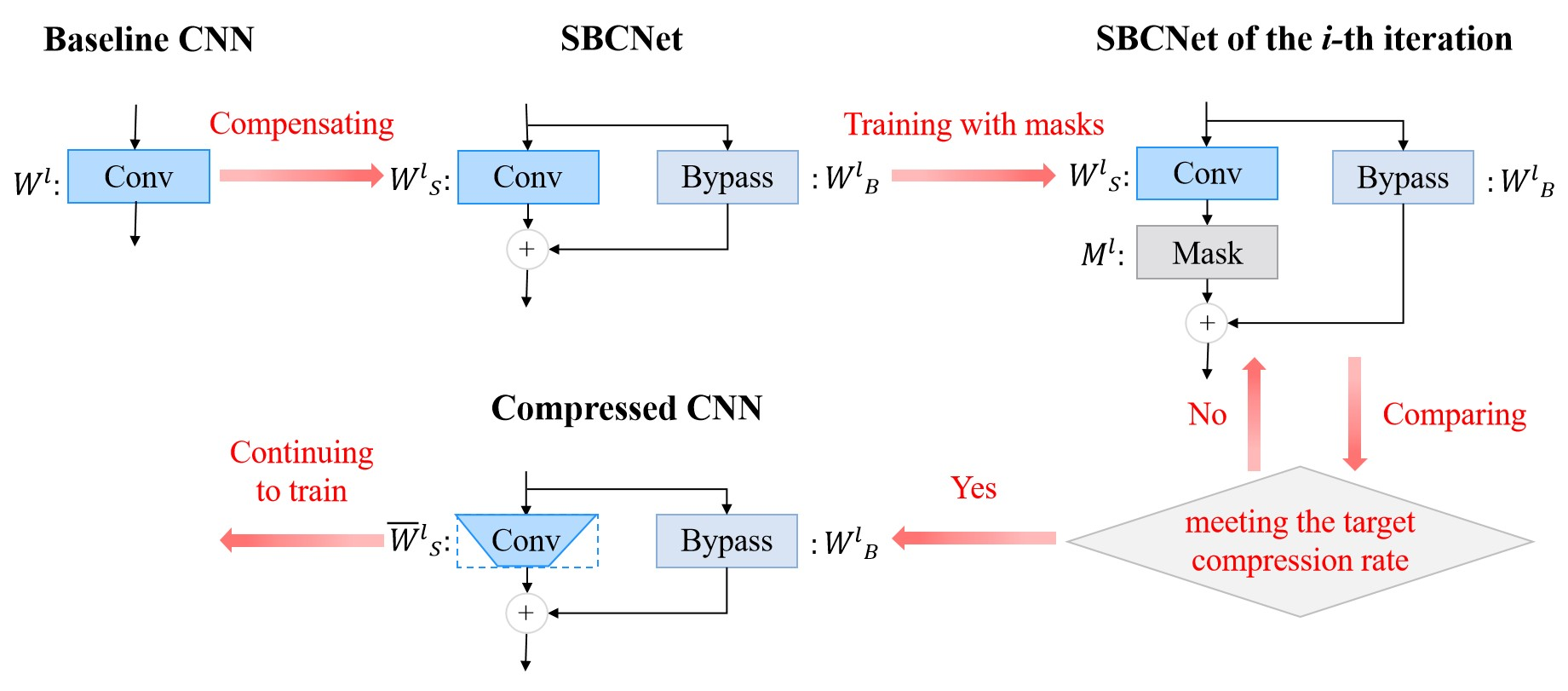}
\caption{The overview of our method. Here, for brevity, batch normalization (BN) layers and nonlinear activation layers are omitted. Before training from scratch, each convolutional layer to be pruned in the baseline CNN is replaced with the sparse module with bypass compensation to build the SBCNet. Then the SBCNet is trained and compressed with masks by the proposed pruning strategy. After meeting the target compression rate, the binary sparse masks are removed, the SBCNet is converted to the compressed network, and then the compressed network is trained for the remaining epochs. }
\label{fig1}
\end{figure*}

Pruning can be divided into unstructured and structured pruning. Unstructured pruning directly removes unimportant weight elements in a filter independently by some importance metrics \cite{ding2019global}, \cite{dong2017learning}, which results in irregular sparsity. It is difficult for irregular structure to leverage existing software and hardware to get an actual speedup. In contrast, structured pruning directly removes the whole unimportant filters (or channels, layers, blocks, etc.)  based on different importance metrics \cite{ding2021resrep}, \cite{lin2020hrank}, and achieves structured sparsity. Therefore, structured pruning, which is able to achieve actual acceleration based on existing software and hardware, has been rapidly developing in recent years. Although existing structured pruning methods, especially filter pruning and channel pruning methods, have achieved inspiring results, there still exists two problems.

Firstly, pruning rate setting is a fundamental problem. The redundancy of different convolutional layers is different, but many pruning methods \cite{he2018soft, he2019filter, zhang2022hierarchical} ignore this and set the same pruning rates for all layers, which may not produce a proper pruning structure. In order to assign different pruning rates to different layers, some works \cite{wang2020pruning}, \cite{liu2017learning} set a fixed global metric threshold, and distinguish important filters based on corresponding importance metrics. However, when given a target compression rate, the threshold need be set by trial and error. What's more, the global threshold is not appropriate for all layers due to the different distribution of metric values in each layer. But manually setting different thresholds for each layer requires more labor costs and causes the suboptimal pruning rates distribution. 

By making metric thresholds of each layer learnable, soft threshold reparameterization (STR)\cite{kusupati2020soft} is able to automatically set thresholds for each layer and obtain better pruning rates distribution. The learning of the thresholds in STR is controlled by the task loss and weight decay regularization. However, by adjusting the weight-decay coefficient, STR cannot explicitly control the thresholds to make the pruned network exactly reach a target compression rate. In order to explicitly control the pruned network to meet the target FLOPs, some other studies \cite{gao2021network}, \cite{gao2020discrete} introduce FLOPs constraints to guide the learning of the pruned structure. To represent the pruned structure, they incorporate an extra learnable parameter for each channel in each layer to form a differentiable gate. By using FLOPs constraints, they can progressively and automatically assign a different pruning rate to each layer during training based on the target FLOPs. However, they introduce too many additional learnable parameters, which makes the optimization of the pruned structure difficult.

Secondly, automatic pruning rate setting cannot explore a high pruning rate for a specific layer. Most pruning methods \cite{gao2021network, shen2022prune, wang2020pruning, nonnenmacher2021sosp} try to identify the least important filters or channels as possible to reduce the loss of information after pruning, and then simply remove them. However, as the pruning rates get higher and higher, the pruned layers lose more and more information and capacity. Especially, when the pruning rates are set very high, these pruned too narrow layers have limited expressive power and probably block the propagation of features and gradients, which may make it difficult for the pruned network to restore performance even after a lot of time of training or fine-tuning.

In order to deal with the above problems, in this paper, we propose a novel Layer Adaptive Progressive Pruning (LAPP) framework, which consists of module design and pruning strategy. An overview of our method is shown in Fig. \ref{fig1}. To deal with the above first problem, i.e., pruning rate setting problem, our pruning strategy takes metric thresholds as learnable parameters and introduces FLOPs constraints to control their updates. Specifically, we use ${\ell}_1$-norm as the filter importance metric. We introduce a learnable threshold for each layer to distinguish the importance between filters, which avoids introducing too many additional learnable parameters. During training, guided by both task loss and FLOPs constraints, the learnable thresholds are dynamically updated to accommodate changes of the filters norm. According to the relative size between the thresholds and the filters norm, our method can progressively automatically distinguish important filters and determine appropriate pruning rates for each layer in each iteration until the target compression rate is met, which greatly saves trial costs.

To address the above second problem, before training starts, by introducing an additional lightweight bypass for the convolutional layer (i.e., the sparse path) to be pruned in the baseline network, we design a  $\textbf{S}$parse module with $\textbf{B}$ypass $\textbf{C}$ompensation, named $\textbf{SBC}$ module, and build the $\textbf{SBCNet}$, as shown in Fig. \ref{fig2}. Then we only progressively prune the sparse paths and keep the bypasses throughout the training to compensate the sparse paths. Therefore, even if the sparse path get very narrow, the SBC module can also have strong expressive power and guarantee the propagation of features and gradients. In this paper, in order to make the lightweight bypass have sufficient expressive power, we finally use the block inspired by MobileNetV2 \cite{sandler2018mobilenetv2} as the bypass. Specifically, the block sequentially contains three layers of small convolutions: $1\times1$ convolution, depthwise convolution, and $1\times1$ convolution.

Last but not least, our pruning strategy is a pruning during training algorithm. Specifically, our pruning strategy only takes up a few epochs in the early stages of training to prune the network gradually, and then continues to train the pruned network that meets the target compression rate for the remaining epochs. Therefore, our pruning strategy is able to reduce substantial pre-training burdens of post-training pruning methods \cite{gao2021network,yu2022topology,alwani2022decore} and generate efficient sub-networks with less accuracy degradation than pruning at initialization methods \cite{wang2020pruning}.

To demonstrate the effect of our method, we conduct experiments on two popular image classification datasets with some representative network structures. Compared to the state-of-the-art methods, LAPP achieves superior performance. For example, we obtain 59.7$\%$ FLOPs reduction by removing 52.8$\%$ parameters without accuracy drop on ResNet-20 compared to baseline. Compared to baseline, 55.6$\%$ of FLOPs of ResNet-18 are reduced with 0.21$\%$ top-1 accuracy increase and 0.40$\%$ top-5 accuracy increase on ImageNet.

Our main contributions are summarized as follows:

(a) Combining learnable thresholds and FLOPs constraints, we design an effective and efficient pruning strategy which can prune the network gradually during training from scratch and determine appropriate pruning rates automatically for each layer.

(b) By introducing an additional bypass for each convolutional layer, our method can compensate the lost information of the pruned layer and guarantee the information flow of the pruned narrow network during training, so as to be able to explore a high pruning rate for a specific layer. 

(c) Our method can be applied to various CNNs, such as VGG \cite{simonyan2014very}, ResNet \cite{he2016deep}, GoogleNet \cite{szegedy2015going}, and DenseNet \cite{huang2017densely}, and achieves state-of-the-art performance on those networks, which shows its effectiveness. 

The rest of the paper is organized as follows. In Section II, we introduce related work. In Section III, we provide a detailed description of our method. Then in Section IV, we conduct experiments on two popular image classification datasets with different types of CNNs and compare the results with those of the state-of-the-art methods. Finally, we conclude in Section V.

\section{RELATED WORK}
\subsection{Pruning for Network Compression}
{\bf{Pruning Pipelines.}} Depending on when pruning is performed, most pruning methods can be divided into three categories\cite{shen2022prune}: 1) post-training pruning, 2) pruning at initialization, 3) pruning during training. The post-training pruning methods \cite{li2016pruning},\cite{ding2021resrep},\cite{gao2021network,yu2022topology,zhuang2018discrimination,alwani2022decore,nonnenmacher2021sosp} often adopt a three-stage pipeline, namely densely pre-training, pruning, and fine-tuning, to compress networks. Because of pre-training over-parameterized networks and fine-tuning pruned networks, post-training pruning methods usually generate efficient sub-networks with plausible accuracy, but result in huge training burdens. Different from post-training pruning methods, pruning at initialization methods \cite{wang2020pruning}, \cite{lee2018snip, hayou2020robust, wang2019picking} are a two-stage training pipeline, namely pruning the randomly initialized network and then training the pruned network from scratch, which greatly avoids the pre-training burdens. However, existing methods still remain unsatisfactory as the resulting pruned networks can be difficult to train \cite{hayou2020robust} and easily suffer significant accuracy degradation.

Pruning during training methods \cite{shen2022prune},\cite{ruan2021dpfps}, \cite{frankle2020linear} aim to combine the benefits of the above two strategies. Pruning during training methods can be further split into two directions \cite{shen2022prune}: regularization-based methods \cite{ruan2021dpfps},\cite{liu2017learning} and sub-ticket selection methods \cite{shen2022prune}, \cite{frankle2020linear}. Similar to post-training pruning methods, regularization-based methods also adopt a three-stage pipeline, namely training with sparsity regularization, pruning, and slightly fine-tuning. Regularization-based methods spend less time on fine-tuning the pruned network, but the training process under sparsity regularization still causes substantial computational burdens. Sub-ticket selection methods select the pruned network structure via some importance metrics when performing pruning, but usually rely on some experience or assumptions to heuristically set the starting points of pruning during normal training. Different from previous sub-ticket selection methods, based on learnable metric thresholds, pruning in our pruning strategy is automatically started during training. Different from previous regularization-based methods, our pruning strategy only takes up a few epochs in the early stages of training to prune the network progressively, and then reduce more training burdens by training the pruned network for the remaining epochs. 

{\bf{Pruning Metrics.}} Pruning metrics are used to measure the importance or redundancy of filters (or channels). Depending on whether they introduce extra learnable parameters, pruning metrics of some representative structured pruning methods can be roughly grouped into two categories. Pruning metrics without extra parameters can be further split into two subcategories: data independent metrics and data dependent metrics.

Some works \cite{li2016pruning,he2018soft,he2020learning,wang2019pruning,he2019filter,kim2022fp,ruan2021dpfps} directly use inherent properties of filters, such as the ${\ell}_1$-norm or ${\ell}_2$-norm values of filters and the relationships among filters, as pruning metrics. Based on the assumption that filters with smaller norms are more unimportant, pruning filters for efficient convnets (PFEC)\cite{li2016pruning} selects and removes the filters with smaller ${\ell}_1$-norm values. Filter pruning via geometric median (FPGM)\cite{he2019filter} points out that the above assumption is not always true, and then turns to pruning the most replaceable filters based on the geometric median of filters in each layer. Considering that similar filters are redundant, different from FPGM, filter pruning with adaptive gradient learning (FP-AGL)\cite{kim2022fp} imposes the redesigned centripetal vectors on filters to converge filters in the clusters to the same point, and finally keeps only one filter per cluster. Some works \cite{liu2017learning,zhuang2020neuron} leverage the scaling factors in batch normalization (BN) layers to judge the importance of channels. Network slimming (NS)\cite{liu2017learning} pushes the scaling factors in BN layers to zero by imposing ${\ell}_1$ regularization during training, and prunes channels with small scaling factors after training. The above metrics are data independent, so these works require low computational costs to determine the pruned filters (or channels). 

Besides, some pruning metrics require to utilize input data to select the pruned filters (or channels). Output feature maps are the convolutional results of input feature maps and filters, so some studies \cite{lin2020hrank,sui2021chip,zhang2022hierarchical} determine the pruned filters by judging the importance or redundancy of feature maps from a set of input data. High rank (Hrank)\cite{lin2020hrank} utilizes the average rank of feature maps to judge the importance of filters. Different from Hrank, channel independence-based filter pruning (CHIP) \cite{sui2021chip} finds redundant feature maps and prunes corresponding filters by measuring the correlations among feature maps in each layer. Considering that filters are updated by gradients from the loss function during training, some studies \cite{zhuang2018discrimination,nonnenmacher2021sosp} count the gradient or second-order derivatives information for each filter (or channel) to evaluate the importance. Discrimination-aware channel pruning (DCP) \cite{zhuang2018discrimination} chooses the most important channels by computing the Frobenius norm of the gradients. Second-order structured pruning (SOSP)\cite{nonnenmacher2021sosp} designs two saliency-based pruning metrics, which use first-order or second-order derivatives information, to delete unimportant filters.

Some works \cite{wang2020pruning,gao2021network,alwani2022decore,elkerdawy2022fire,yu2022topology} introduce extra learnable parameters and optimize these parameters leveraging reinforcement learning algorithms or gradient based methods, to generate a saliency score or binary decision for each channel (or filter). Deep compression with reinforcement learning (DECORE)\cite{alwani2022decore} assigns an agent with only one parameter to each channel and learns which channels to be pruned based on reinforcement learning. For each input sample, fire together wire together (FTWT)\cite{elkerdawy2022fire} utilizes a well-trained binary mask predictor head in each layer to dynamically predict $k$ activated filters.

\subsection{Combining Pruning and Other Techniques for Network Compression}
{\bf{Pruning and Low-rank Tensor Decomposition.}} Low-rank tensor decomposition (LTD)\cite{phan2020stable} decomposes the low-rank weight tensor of the convolution into a sequence of tensors which contain much fewer parameters and computations. Since pruning and LTD reduce redundancy in parameters from different angles, some recent efforts \cite{guo2019compressing, li2021towards,li2020group} combine them together to pursue better network compression performance. Li et al. \cite{li2020group} proposed to attach sparsity-inducing matrices to normal convolutions and apply group sparsity constraints on them, to hinge filter pruning and decomposition (Hinge). By simultaneously dealing with the sparsity and low-rankness in weights, collaborative compression (CC)\cite{li2021towards} combines channel pruning and tensor decomposition to accelerate CNNs.

{\bf{Pruning and Low-bit Quantization.}} Low-bit quantization\cite{sharma2021generalized,li2021residual,duan2022differential} can effectively reduce computation and storage cost by reducing the number of bits used to represent the network weights or activations. Quantization is orthogonal to pruning, so some works \cite{wang2020differentiable,van2020bayesian,wang2022fast} jointly perform them to achieve better network acceleration. Wang et al.\cite{wang2020differentiable} proposed to integrate structured pruning with mixed-bit precision quantization via a joint gradient-based optimization scheme. Integer-only Discrete Flows (IODF) \cite{wang2022fast} performs efficient invertible transformations by utilizing integer-only arithmetic based on 8-bit quantization and introducing learnable binary gates to remove redundant filters during inference.

{\bf{Pruning and Knowledge Distillation.}} Knowledge distillation (KD)\cite{hinton2015distilling} transfers the knowledge from a large teacher network to a small student network. Recently, some works exploit KD to enhance the learning ability of the pruned networks \cite{li2022learning,zou2022dreaming} or exploit pruning to enhances the KD quality \cite{liu2021content,park2022prune}. In order to preserve the performance of the pruned model, Zou et al.\cite{zou2022dreaming} proposed to distill the pruned model to fit the pre-trained model by utilizing reconstructed degraded images. In order to boost the performance of KD, Park and No\cite{park2022prune} proposed to prune the teacher network first to make it more student-friendly and then distill it to the student.

The above works further show that pruning and other techniques are complementary. Therefore, combining them can improve the compression performance further.

\begin{figure}[!t]
	\centering
	\includegraphics[width=3.1in]{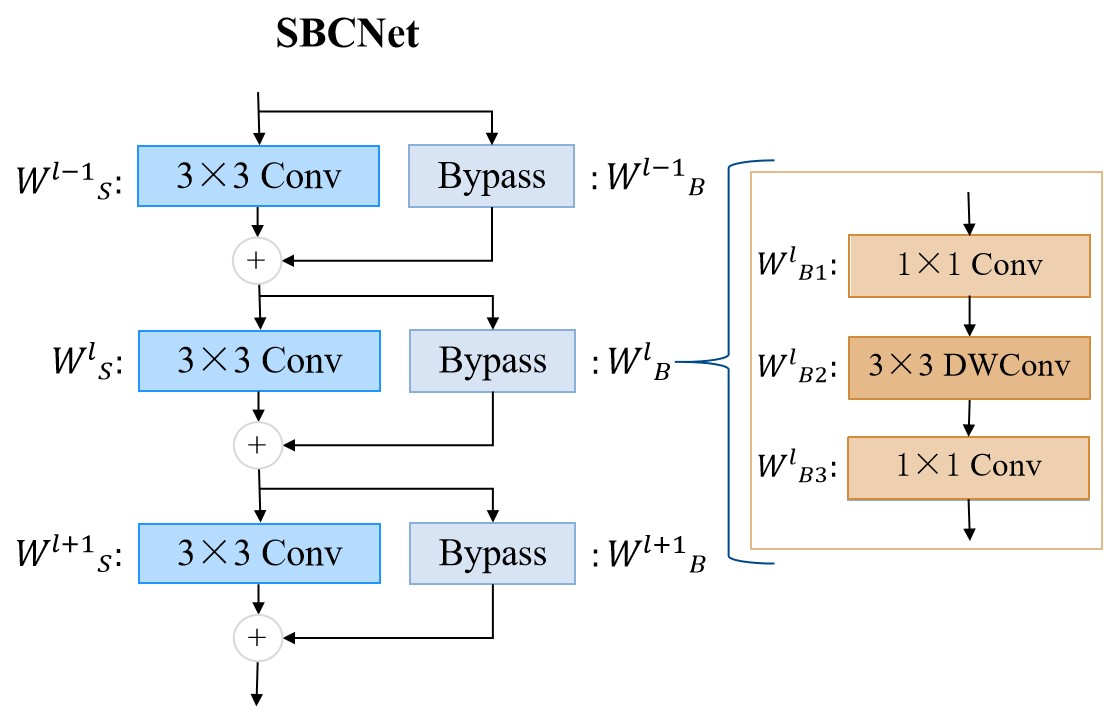}
	\caption{The $3\times3$ SBC modules in the SBCNet. Here the bypass is the block inspired by MobileNetV2 \cite{sandler2018mobilenetv2}. "DWConv" denotes a depthwise convolution, and all BN layers and nonlinear activation layers are omitted.}
	\label{fig2}
\end{figure}

\section{PROPOSED METHOD}
In this section, we introduce the framework and compression process of our proposed method.
\subsection{Preliminaries}
For the $l$-th layer of a CNN, we denote the input feature map as $X^l\in \mathbb{R}^{c^{l-1}\times h^{l-1}\times w^{l-1}}$, the output feature map as $Y^l\in \mathbb{R}^{c^{l}\times h^{l}\times w^{l}}$, and weight tensor of the filters as $W^l\in \mathbb{R}^{c^{l}\times c^{l-1}\times k^{l}\times k^{l}}$, where $c^{l-1}$ and  $c^{l}$ are the number of input channels and filters (output channels), respectively, and $k^{l}$ is the kernel size of the filters. The output of the $l$-th convolutional layer can be represented as:
\begin{equation}\label{convolution operation}
Y^l = W^l \otimes X^l,
\end{equation}
where $\otimes$ denotes the convolution operation. 
 
{\bf{Structured Pruning.}} Both filter pruning and channel pruning seek a compact approximated representation $\overline{W}^l$. In filter pruning, each filter is regarded as a compression unit and the pruning function is defined as:
\begin{equation}\label{filter pruning}
\overline{W}^l [i,:,:,:] = W^l [K_f^l[i],:,:,:]
\end{equation}
where $\overline{W}^l\in \mathbb{R}^{n^{l}\times c^{l-1}\times k^{l}\times k^{l}}$ and $K_f^l$ denotes an list consisting of the indexes of the kept filters in $W^l$. Here $n^{l}$ is the number of the kept filters. In channel pruning, each input channel is regarded as a compression unit and the pruning function is defined as:
\begin{equation}\label{channel pruning}
\overline{W}^l [:,i,:,:] = W^l [:,K_c^l[i],:,:]
\end{equation}
where $\overline{W}^l\in \mathbb{R}^{c^{l}\times n^{l-1}\times k^{l}\times k^{l}}$ and $K_c^l$ denotes an list consisting of the indexes of the kept input channels in $W^l$. Here $n^{l-1}$ is the number of the kept input channels.

In this paper, we introduce our pruning method from the perspective of filter pruning, but note that it is also suitable for channel pruning. In the following, for brevity, we replace $W^l [i,:,:,:]$ with $W^l [i]$.

\begin{figure*}[!t]
	\centering
	\includegraphics[width=6.0in]{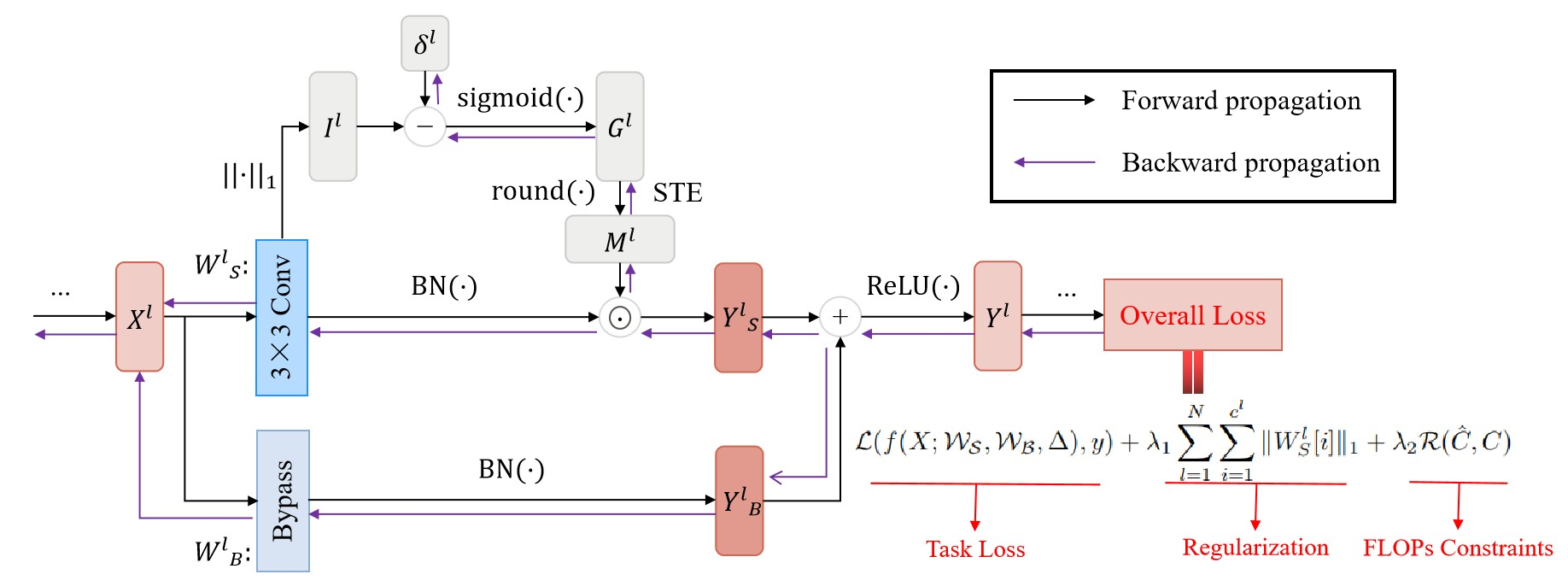}
	\caption{The pipeline of our pruning strategy for the $l$-th SBC module. Here STE denotes straight-through estimator. For the $l$-th SBC module, in the process of forward propagation, ${\ell}_1$-norm is used to calcuate the importance score $I^l$ of the filters weight $W_S^l$ in the sparse path, and then the difference between the importance score $I^l$ and the learnable threshold ${\delta}^l$ is input into the function $\text{round}(\text{sigmoid}(\cdot))$ to get the binary mask $M^l$. The output of the input $X^l$ through the convolutional layer is multiplied by the mask $M^l$ to get the output $Y_S^l$ in the sparse path. Then the output $Y_S^l$ of the sparse path and the output $Y_B^l$ of the bypass are summed to get the final output $Y^l$. In the process of backward propagation, the learnable threshold ${\delta}^l$ and filters weight $W_S^l$ and $W_B^l$ are updated by the guidance from the gradients of the overall loss. Note that in order to stabilize the compression training and not to interfere with the learning of $W_S^l$, gradient propagation from $G^l$ to $W_S^l$ is truncated, and gradients from $G^l$ are only utilized to guide the update of the threshold ${\delta}^l$.}. 
	\label{fig3}
\end{figure*}

\subsection{Module Design}
We firstly design a sparse module with bypass compensation, named SBC module, and then replace each convolutional layer to be pruned in the baseline network with the SBC module to build the SBCNet. The SBC module consists of a sparse path (i.e., original convolutional layer) and a lightweight bypass, and the output of the SBC module is the sum of the outputs of the original convolutional layer and the lightweight bypass. Therefore, the SBCNet adds a few additional burdens compared to the baseline network. An example for the $3\times3$ SBC module is shown in Fig.\ref{fig2}. In this paper, we use the block inspired by MobileNetV2 \cite{sandler2018mobilenetv2} as the bypass to make it have sufficient expressive power. Besides, the structure of the block can be obtained by applying LTD to a regular convolution \cite{kossaifi2020factorized}, so using the block as the bypass can also make the SBC module exploit the complementarity of pruning and LTD to improve the performance of our method further. For the $l$-th SBC module, the output is represented as:
\begin{equation}\label{convolution operation using new convolutional module}
Y^l = Y^l_S + Y^l_B = W^l_S \otimes X^l + W^l_{B3}\otimes(W^l_{B2}\otimes (W^l_{B1} \otimes X^l)),
\end{equation}
where $W^l_S\in \mathbb{R}^{c^{l}\times c^{l-1}\times k^{l}\times k^{l}}$ denotes the weight tensor of the sparse path, while $W^l_{B1}\in \mathbb{R}^{{d}^l \times c^{l-1} \times 1 \times 1}$, $W^l_{B2}\in \mathbb{R}^{{d}^l \times 1 \times k^{l} \times k^{l}}$, and $W^l_{B3} \in \mathbb{R}^{c^{l}\times {d}^l \times 1 \times 1}$ denote the weight tensors of the lightweight bypass.

For the sparse path, $W^l_S$ is randomly initialized before training, and is made sparse with a mask by the proposed pruning strategy during training until the target compression rate is met. Detailed explanations of the proposed pruning strategy are given in the following sub-section.

For the lightweight bypass, the three lightweight convolutions are kept throughout the training to compensate the sparse path. Suppose the stride of the convolutions in the $l$-th module is 1, $c^l$ and $c^{l-1}$ are equal, and the pruning rate adaptively assigned by our pruning strategy in the sparse path is $p^l$. For brevity, we omit superscript. After pruning, the FLOPs of the bypass is $hw(2cd+dk^2)$, and the FLOPs of the SBC module is  $hw(2cd+dk^2+(1-p)c^2k^2)$. In general, $k^2 \ll c$ and $d \le c$, we can ignore the item $dk^2$. Thus the ratio of their FLOPs is approximately $2 \over {2+((1-p)ck^2)/d}$. When $p$ is a small value close to 0, the pruned sparse path is dominant in the SBC module, and the lightweight bypass only brings relatively few calculations. When $p$ is a medium value, the bypass compensates the lost information and capacity of the pruned sparse path with relatively small computational costs. When $p$ is a large value close to 1, the sparse path is pruned so narrow that it blocks the forward propagation of features and back propagation of gradients, which generally results in a drastic decrease in network performance. However, the bypass helps guarantee the propagation of features and gradients during training so as to make the network converge normally. 

In the extreme case, where $p$ is 1, the entire sparse path can be smoothly removed during pruning due to the bypass, which achieves approximately ${ck^2 \over {2d}}\times$ acceleration, and the bypass makes the module still maintain relatively strong expression power. However, this does not mean that it is more effective to directly build a compact network using bypasses instead of SBC modules before training starts. Because as mentioned above, some modules need remain some channels of sparse paths to compensate their lightweight bypasses so as to maintain sufficient capacity. The presence of sparse paths can increase the diversity of features. Besides, they can act as shortcuts to facilitate the flow of information, and pruning dense SBCNet during training can help search a better sub-network.

What's more, by adjusting the hyperparameter ${d}^l$, the proportion of FLOPs of the bypass to the FLOPs of the pruned SBC module can be roughly controlled. Therefore, by setting the appropriate value for ${d}^l$, we can achieve the compression goal and make the bypass generate sufficient compensation for the sparse path so as to achieve better compression performance. In this paper, when the global target compression rate is set to a large value, ${d}^l$ is empirically set directly to $c^{l}$; when the global target compression rate is set to a very small value, ${d}^l$ is reduced to $0.5c^{l}$. 

\subsection{Pruning Strategy}
The post-training pruning is a common used pruning strategy. However, its three-stage pipeline, namely densely pre-training, pruning, and fine-tuning, results in nearly twice training time. Due to its inefficiency, one intuitive solution is pruning at initialization. Although pruning at initialization makes the training more efficient as it only trains the pruned network, the pruned networks can be difficult to train and easily suffer significant accuracy degradation. Therefore, pruning during training, which can reduce negative effects of post-training pruning and pruning at initialization, probably finds a trade-off between training efficiency and final accuracy \cite{shen2022prune}. So our pruning strategy also gradually prunes the sparse paths during initial training of a few epochs, and  exploits the remaining large number of epochs to train the pruned network to converge to improve performance. The pipeline of our pruning strategy for the $l$-th module is shown in Fig.\ref{fig3}.

Specifically, in this paper, for simplicity, we use ${\ell}_1$-norm as the filter importance metric. Thus the filter importance score ${I}^l \in \mathbb{R}^{c^l}$ of the sparse path is represented as:
\begin{equation}\label{filter importance score}
{I}^l = [ \Vert W^l_S [1]\Vert_1,\Vert W^l_S [2]\Vert_1,\cdots,\Vert W^l_S [c^l]\Vert_1]^{T},
\end{equation}
in which each element represents the importance of the corresponding filter. Different from PFEC\cite{li2016pruning} which directly manually sets the pruning rates for each layer, we use pruning thresholds to determine the pruning rates for each layer. Suppose the threshold for the $l$-th module is ${\delta}^l$, then the pruning mask ${M}^l \in \mathbb{R}^{c^l}$ is obtained according to ${I}^l$:
\begin{equation}\label{pruning masks}
{M}^l [i] = \begin{cases} 1,& {I}^l_{i} \geq {\delta}^l, \\ 0,& {I}^l_{i}<{\delta}^l,  \end{cases}
\end{equation}
where ${I}^l_{i}$ denotes the $i$-th element of ${I}^l$. By summing the mask ${M}^l$, the number of the kept filters can be obtained, namely ${n}^l$, and the pruning rate of the sparse path is  $p^l = 1-n^l/c^l$. The output of the sparse path is then reformulated as follows:
\begin{equation}\label{convolution operation when using pruning masks}
Y^l_S = {M}^l \odot (W^l_S \otimes X^l),
\end{equation}
where $\odot$ is the element-wise product between the column vector on the left and the spatial dimensions of the three-dimensional tensor on the right.

Since different layers' parameters distribution are different, setting same pruning thresholds for all layers may not be optimal. However, manually setting different thresholds for each layer requires lots of labor costs and may also cause the suboptimal pruning rates distribution. Inspired by the idea of applying learnable thresholds to filter pruning proposed by Kusupati et al. \cite{kusupati2020soft}, we combine the simple ${\ell}_1$-norm metric with learnable thresholds by introducing an additional learnable parameter for each sparse path of the network.

From Eq.(\ref{convolution operation when using pruning masks}), ${M}^l$ is directly involved in the forward calculation of the network and can receive gradients from the objective function during backpropagation. However, according to Eq.(\ref{pruning masks}), the threshold ${\delta}^l$ is non-differentiable. Therefore, we change ${M}^l$ into the soft pruning mask ${G}^l$ by introducing $\text{sigmoid}(\cdot)$ function:
\begin{equation}\label{soft pruning masks using sigmoid function}
{G}^l = \text{sigmoid}({I}^l-{\delta}^l) = {1 \over {1+e^{-({I}^l-{\delta}^l)}}}.
\end{equation}
Here each element of ${G}^l$, i.e., ${G}^l[i]$, is continuous and its value ranges from 0 to 1. Following the works \cite{gao2021network},\cite{elkerdawy2022fire}, we further round ${G}^l[i]$ to 0 or 1, to precisely generate sub-networks during training, as follows:
\begin{equation}\label{hard pruning masks using indicator function}
{M}^l[i] = \mathds{1}_{\geq 0.5}({G}^l[i]), \,\forall{i\in[1,c^l]}.
\end{equation}
When the value of ${G}^l[i]$ is greater than or equal to 0.5, i.e., when the corresponding filter importance score ${I}^l_{i}$ is greater than or equal to the threshold ${\delta}^l$, the value of the corresponding mask ${M}^l[i]$ is 1, otherwise, it is 0. However, the indicator function $\mathds{1}_{ \geq 0.5}(\cdot)$ is still non-differentiable. To tackle this problem, we use straight-through estimator (STE) \cite{bengio2013estimating} to calculate the gradients from ${M}^l$ to ${G}^l$ during back propagation, making the update of ${\delta}^l$ feasible.

\begin{algorithm}[H]
	\caption{Overview of the LAPP method.}\label{alg:alg1}
	\begin{algorithmic}
		\STATE 
		\STATE {\textbf{Input:}} $\text{CNN with } N \text{ convolutional layers to be pruned}; \text{ t-}$
		\STATE $\text{otal epochs } E; \text{ training set }D; \text{ total FLOPs }T_{total}\text{ of the}$
		\STATE $\text{CNN; target compression rate }C;$
		\STATE {\textbf{Output:}} $\text{Compact CNN satisfying the target compression }$
		\STATE $\text{rate }C \text{ and its optimal weight values } \mathcal{\overline{W}_S} \text{ and } \mathcal{W_B};$
		\STATE {\text{ 1:}} $\text{Build the SBCNet with }\mathcal{W_S} \text{ and } \mathcal{W_B} \text{ based on Eq.(\ref{convolution operation using new convolutional module});}$
		\STATE {\text{ 2:}} $\text{Introduce learnable thresholds} \text{ for each sparse path};$
		\STATE {\text{ 3:}} $\text{Enforce epoch status}\in\text{\{``prune'', ``train''\}};$
		\STATE {\text{ 4:}} $\text{epoch status is set as ``prune''};$
		\STATE {\text{ 5:}} $\textbf{for}\text{ epoch } t = 1, 2, . . . , E \textbf{ do}$
		\STATE {\text{ 6:}}\hspace{0.5cm} $\textbf{for}\text{ a mini-batch (}X,y\text{) in }D\textbf{ do}$
		\STATE {\text{ 7:}}\hspace{1.0cm}$\textbf{if} \text{ epoch status is ``prune'' } \textbf{then}$
		\STATE {\text{ 8:}}\hspace{1.5cm}$\textbf{for}\text{ layer } l = 1, 2, . . . , N \textbf{ do}$
		\STATE {\text{ 9:}}\hspace{2.05cm}$\text{Get }{I}^l \text{ by Eq.(\ref{filter importance score})};$
		\STATE {\text{10:}}\hspace{2.0cm}$\text{Get }{M}^l \text{ by Eq.(\ref{soft pruning masks using sigmoid function}) and Eq.(\ref{hard pruning masks using indicator function})};$
		\STATE {\text{11:}}\hspace{2.0cm}$\text{Get } Y^l \text{ by Eq.(\ref{convolution operation when using pruning masks}) and Eq.(\ref{convolution operation using new convolutional module})};$
		\STATE {\text{12:}}\hspace{1.5cm}$\textbf{end for}$
		\STATE {\text{13:}}\hspace{1.5cm}$\text{Calculate }T_{kept} \text{ and }\hat{C};$
		\STATE {\text{14:}}\hspace{1.5cm}$\textbf{if } \hat{C} \neq C \textbf{ then}$
		\STATE {\text{15:}}\hspace{2.0cm}$\text{Update }\mathcal{W_S}, \mathcal{W_B} \text{ and }\Delta \text{ with SGD};$
		\STATE {\text{16:}}\hspace{1.5cm}$\textbf{else}$
		\STATE {\text{17:}}\hspace{2.0cm}$\text{Remove the masks and prune, get } \mathcal{\overline{W}_S};$
		\STATE {\text{18:}}\hspace{2.0cm}$\text{Set epoch status as ``train''};$
		\STATE {\text{19:}}\hspace{1.5cm}$\textbf{end if}$
		\STATE {\text{20:}}\hspace{1.0cm}$\textbf{else}$
		\STATE {\text{21:}}\hspace{1.5cm}$\text{Get } Y^l\text{ for each layer } l\text{ by Eq.(\ref{convolution output of the compressed network})};$
		\STATE {\text{22:}}\hspace{1.5cm}$\text{Update }\mathcal{\overline{W}_S} \text{ and } \mathcal{W_B} \text{ with SGD};$ 
		\STATE {\text{23:}}\hspace{1.0cm}$\textbf{end if}$
		\STATE {\text{24:}}\hspace{0.5cm} $\textbf{end for}$
		\STATE {\text{25:}} $\textbf{end for}$
		\STATE {\text{26:}}{\textbf{Return:}} $\text{Compact CNN with weights }\mathcal{\overline{W}_S} \text{ and } \mathcal{W_B};$
	\end{algorithmic}
	\label{alg1}
\end{algorithm}

To make the filters be more distinguishable, we apply sparse regularization to the filter weights of the sparse paths. Consistent with the ${\ell}_1$-norm importance metric, we use ${\ell}_1$ regularization, and obtain the network optimization problem:
\begin{equation}\label{optimization problem with sparse regularization}
\mathop{\min}_{\mathcal{W_S},\mathcal{W_B},\Delta} {\mathcal{L}(f(X;\mathcal{W_S},\mathcal{W_B},\Delta),y)+\lambda\sum_{l=1}^N\sum_{i=1}^{c^l}\Vert W^l_S [i]\Vert_1},
\end{equation}
where $\mathcal{L}(\cdot)$ denotes the cross-entropy loss, and $\mathcal{W_S}=\{W^1_S,W^2_S,\cdots,W^N_S\}$ and $\mathcal{W_B}=\{W^1_B,W^2_B,\cdots,W^N_B\}$ refer to the parameters in the $N$ sparse modules of the SBCNet $f(\cdot;\mathcal{W_S},\mathcal{W_B},\Delta)$. Here $X$ and $y$ represent the input samples and corresponding labels respectively, ${\Delta}$ represents the set of all learnable metric thresholds of the sparse paths of the SBCNet, $\lambda$ is the coefficient of ${\ell}_1$ regularization term, and the weight decay regularization term is ignored for brevity. Therefore, the final overall sparsity of the sparse paths of the network is controlled by the coefficient of ${\ell}_1$ regularization and the initial value of the thresholds in ${\Delta}$. However, adjusting these hyperparameters by trial and error to achieve the target compression rate is undoubtedly tedious and complicated. Therefore, to explicitly control the sparsity of the network, i.e., making the final sub-network reach given target FLOPs, we introduce FLOPs constraints to control the update of the thresholds in ${\Delta}$. Previously FLOPs constraints and metric thresholds are often used separately, while we attempt to combine learnable thresholds and FLOPs constraints to automatically determine the appropriate pruning rates for each layer based on the target compression rate during training. 

Different from STR \cite{kusupati2020soft} which applys the $\text{sigmoid}(\cdot)$ function to the learnable threshold, we apply it to the difference between the importance score $I^l$ and the learnable threshold ${\delta}^l$, i.e., Eq.(\ref{soft pruning masks using sigmoid function}), to control the value range of the soft mask ${G}^l$ from 0 to 1. Here, $\text{ReLU}(\cdot)$ activation function used in STR is unnecessary. Then we further introduce the indicator function and STE, i.e., Eq.(\ref{hard pruning masks using indicator function}), so as to obtain the learnable binary pruning masks $\mathcal{M}=\{M^1,M^2,\cdots,M^N\}$ which characterize the sub-network structure. Therefore, we can precisely calculate the FLOPs $T_{kept}$ of the sub-network at a certain iteration by the binary masks $\mathcal{M}$. Suppose the total FLOPs of the baseline network is $T_{total}$, we can obtain the compression rate $\hat{C} = T_{kept}/T_{total}$. Note that since the size of the SBCNet is larger than the baseline, the initial value of $\hat{C}$ is greater than 1 before training starts. In order to control the thresholds to make the sub-network smoothly and accurately reach the given target FLOPs $CT_{total}$, where $C$ is the target compression rate, we design a regularization term to constrain the FLOPs, as follows:
\begin{equation}\label{FLOPs constraints}
\mathcal{R}(\hat{C},C) = ({\hat{C} \over C} - 1)^2.
\end{equation}
Then the optimization problem of the network is reformulated as:
\begin{equation}\label{optimization problem with sparse regularization and FLOPs constraints}
\begin{aligned}
\mathop{\min}_{\mathcal{W_S},\mathcal{W_B},\Delta} &{\mathcal{L}(f(X;\mathcal{W_S},\mathcal{W_B},\Delta),y)}\\
&+{\lambda}_1\sum_{l=1}^N\sum_{i=1}^{c^l}\Vert W^l_S [i]\Vert_1 +{\lambda}_2\mathcal{R}(\hat{C},C),
\end{aligned}
\end{equation}
where ${\lambda}_1$ and ${\lambda}_2$ are coefficients of the ${\ell}_1$ regularization term and FLOPs regularization term, respectively. In order to control the thresholds  $\delta$ more flexibly and quickly to compress the network to the target compression rate $C$, weight decay regularization is no longer applied to the thresholds in ${\Delta}$. By the masks $\mathcal{M}$, the learning of the thresholds are guided jointly by the task loss $\mathcal{L}(f(X;\mathcal{W_S},\mathcal{W_B},\Delta),y)$ and FLOPs regularization term $\mathcal{R}(\hat{C},C)$. Specifically, the thresholds are updated by the guidance from the gradients of $\mathcal{R}(\hat{C},C)$ to drive masks sparse until the resulting sub-network satisfies the target FLOPs, while they are updated by the guidance from the gradients of $\mathcal{L}(f(X;\mathcal{W_S},\mathcal{W_B},\Delta),y)$ to suppress mask sparsity and thus improve the accuracy of the resulting sub-network. Therefore, during initial training of a few epochs, the thresholds, learned in the competition to maintain accuracy and compress the SBCNet to the target FLOPs, result in a relatively better sub-network. 

After the target compression rate is met, we remove the binary sparse masks on the sparse paths, and convert the SBCNet to the compressed network. According to ${M}^l$, we can obtain the index list $K_f^l$ and the pruned weight tensor $\overline{W}^l_S\in \mathbb{R}^{n^{l}\times c^{l-1}\times k^{l}\times k^{l}}$ in the $l$-th sparse path. And $Y^l_S$ is the output of the $l$-th sparse path. So after pruning, the output of the $l$-th SBC module is represented as:
\begin{equation}\label{convolution output of the compressed network}
Y^l [i] = \begin{cases} Y^l_S [\text{Pos}(i)]+Y^l_B [i],& i\in K_f^l, \\ Y^l_B [i],& i\notin K_f^l,  \end{cases}
\end{equation}
where $i\in[1,c^l]$ and function $\text{Pos}(\cdot)$ returns the position of the element $i$ in $K_f^l$. Then, we continue to train the compressed network for the remaining epochs. Finally we can get a sub-network with weights $\mathcal{\overline{W}_S}=\{\overline{W}^1_S,\overline{W}^2_S,\cdots,\overline{W}^N_S\}$ and $\mathcal{W_B}$ that satisfies the target compression rate without compromising performance.

The whole process is summarized in Algorithm \ref{alg:alg1}. Note that although in this paper we take ${\ell}_1$-norm as the filter importance metric due to its simplicity, our pruning strategy can be also combined with some other importance metrics of existing methods.

\section{EXPERIMENTS}
In this section, our method LAPP is evaluated on two benchmark image classification datasets: CIFAR-10 \cite{krizhevsky2009learning} and ImageNet \cite{russakovsky2015imagenet}. We compare our method with some state-of-the-art CNN compression methods, including PFEC\cite{li2016pruning}, DECORE\cite{alwani2022decore}, Hinge\cite{li2020group}, pruning from scratch (PFS)\cite{wang2020pruning}, CC\cite{li2021towards}, NS\cite{liu2017learning}, Hrank\cite{lin2020hrank}, FTWT\cite{elkerdawy2022fire}, SOSP\cite{nonnenmacher2021sosp}, hierarchical pruning via shape-edge representation of feature maps (HPSE)\cite{zhang2022hierarchical}, dynamic and progressive filter pruning scheme (DPFPS)\cite{ruan2021dpfps}, CHIP\cite{sui2021chip}, learning filter pruning criteria (LFPC)\cite{he2020learning}, network pruning via performance maximization (NPPM)\cite{gao2021network}, FP-AGL\cite{kim2022fp}, soft filter pruning(SFP)\cite{he2018soft}, FPGM\cite{he2019filter}, pruning using graph neural networks and reinforcement learning (GNN-RL)\cite{yu2022topology}, and DCP\cite{zhuang2018discrimination}. All the compared methods applied on the mainstream CNN models, including VGG \cite{simonyan2014very}, ResNet \cite{he2016deep}, GoogleNet \cite{szegedy2015going}, and DenseNet \cite{huang2017densely}. We conduct comprehensive experiments with Pytorch \cite{paszke2019pytorch} to show that LAPP is superior or comparable to the above methods.

\subsection{Datasets and Experimental Settings}
{\bf{Datasets.}} The CIFAR-10 dataset contains 60,000 color images with the size of $32 \times 32$ in 10 classes, and each class consists of 6,000 images, 5,000 of which are in the training set and the rest are in the test set. Following \cite{lin2020hrank}, for the training set, we pad 4 pixels with zeros on each side of a image, then randomly crop a $32\times32$ sample from the padded image, and finally randomly flip the cropped image horizontally with a probability of 0.5. For the test set, we directly use the original $32 \times 32$ images. Both training and test images are normalized using channel means and standard deviations. The ImageNet dataset is a large-scale image classification dataset consisting of approximately 1.28 million training images and 50,000 validation images from 1,000 classes. We adopt the same data augmentation strategy as CC \cite{li2021towards}.

\begin{table}
	\begin{center}
		\caption{Compression results of VGG-16 on CIFAR-10.\label{vgg16}}
		\begin{tabular}{m{1.7cm}|c|c|c}
			\hline
			Method & Top-1(\%) &   FLOPs ↓(\%)  & Params ↓(\%)  \\
			\hline\hline
			Baseline & 93.96 & 0.0 & 0.0  \\
			PFEC\cite{li2016pruning} & 93.40 & 34.2 & 64.0  \\
			DECORE\cite{alwani2022decore} & 94.02 & 35.3 & 63.0 \\
			Hinge\cite{li2020group}&94.02&39.1&80.1\\
			PFS\cite{wang2020pruning}&93.63±0.06&50.0&—\\
			CC\cite{li2021towards}&94.15&50.8&65.9\\
			NS\cite{liu2017learning}&93.80&51.0&88.5\\
			Hrank\cite{lin2020hrank}&93.43&53.5&82.9\\
			\textbf{Ours}($C$=0.46)&\textbf{94.32±0.20}&\textbf{53.6}&73.0\\
			\hdashline
			FTWT\cite{elkerdawy2022fire}&93.73&56.0&—\\
			SOSP\cite{nonnenmacher2021sosp}&93.73±0.16&57.7&87.3\\
			CC\cite{li2021towards}&94.09&60.7&72.7\\
			DECORE\cite{alwani2022decore}&93.56&64.8&89.0\\
			Hrank\cite{lin2020hrank}&92.34&65.3&82.1\\
			\textbf{Ours}($C$=0.34)&\textbf{94.36±0.08}&\textbf{65.7}&74.3\\
			\hdashline
			HPSE\cite{zhang2022hierarchical}&93.50&66.1&—\\
			DPFPS\cite{ruan2021dpfps}&93.52±0.15&70.9&93.3\\
			\textbf{Ours}($C$=0.26)&\textbf{93.79±0.16}&\textbf{73.8}&85.2\\
			\hdashline
			Hrank\cite{lin2020hrank}&91.23&76.5&92.0\\
			CHIP\cite{sui2021chip}&93.18&78.6&87.3\\
			DECORE\cite{alwani2022decore}&92.44&81.5	&96.6\\
			\textbf{Ours}($C$=0.18)&\textbf{93.54±0.23}&\textbf{81.9}&86.1\\
			HPSE\cite{zhang2022hierarchical}&92.49&82.0&—\\
			\hline
		\end{tabular}
	\end{center}
\end{table}

{\bf{Experimental Setting.}} On CIFAR-10, the baselines are trained using the same training configurations as the work of Li et al. \cite{li2021towards}, i.e., the initial learning rate is set to 0.1 and is multiplied by 0.1 at 50$\%$ and 75$\%$ of the total 300 epochs. Note that due to the presence of a fine-tuning stage, most of the compared compression methods train more than 400 epochs. So for fair comparison and making the compressed network fully trained, we train the networks except DenseNet from scratch with LAPP for 400 epochs via Stochastic Gradient Descent algorithm (SGD) with momentum 0.9 and an initial learning rate of 0.1. The mini-batch size and weight decay are set to 128 and 0.0001, respectively. Except the number of epochs, our training configurations are the same as fine-tuning configurations of CC \cite{li2021towards}. To achieve comparable performance, we train DenseNet from scratch with LAPP for 600 epochs which is same as the total number of epochs of CC \cite{li2021towards}. On ImageNet, the networks are trained from scratch with LAPP for 120 epochs with the initial learning rate of 0.1. The learning rate is divided by 10 at epoch 30, 60 and 90, and the mini-batch size is set to 256. All the learnable thresholds are empirically initialized to 0, and the coefficient ${\lambda}_2$ of the FLOPs constraints term is set to 1.0 for all networks except ResNet34. The coefficient ${\lambda}_1$ of ${\ell}_1$ regularization term is set to 3e-5 for ResNet-20 and 2e-5 for other networks on CIFAR-10. On ImageNet,  ${\lambda}_1$ is set to 1e-5, and ${\lambda}_2$ is set to 0.5 for ResNet-34. All experiments are repeated for three times and the average results are reported. Note that in all subsequent tables, FLOPs ↓ and Params ↓ represent the reduction of FLOPs and parameters of the compressed network compared to the Baseline network, respectively.

\subsection{Experimental Results on CIFAR-10}
We conduct experiments on the CIFAR-10 dataset with several popular CNNs, including ResNet-56/110/20/32, VGG-16, DenseNet-40 and GoogLeNet. Following \cite{lin2020hrank}, VGG-16 is a modified version of the original and the output of the original GoogLeNet is changed to fit the class number in CIFAR-10.

\begin{table}
	\begin{center}
		\caption{Compression results of ResNet-56/110 on CIFAR-10.\label{resnet56/110}}
		\begin{tabular}{m{1.7cm}|c|c|c}
			\hline
			Method & Top-1(\%) &   FLOPs ↓(\%)  & Params ↓(\%)  \\
			\hline\hline
			ResNet-56&93.33&0.0&0.0\\
			LFPC\cite{he2020learning}&93.72±0.29&47.1&—\\
			DECORE\cite{alwani2022decore}&93.26&49.9&49.0\\
			PFS\cite{wang2020pruning}&93.05±0.19&50.0&—\\
			Hrank\cite{lin2020hrank}&93.17&50.0&42.4\\
			NPPM\cite{gao2021network}&93.40&50.0&—\\
			Hinge\cite{li2020group}&93.69&50.0&48.7\\
			CC\cite{li2021towards}&93.64&52.0&48.2\\
			FP-AGL\cite{kim2022fp}&93.69&52.5&—\\
			\textbf{Ours}($C$=0.47)&\textbf{93.72±0.16}&\textbf{52.5}&40.0\\
			\hdashline
			SFP\cite{he2018soft}&93.35±0.31&52.6&—\\
			FPGM\cite{he2019filter}&93.49±0.13&52.6&—\\
			DPFPS\cite{ruan2021dpfps}&93.20±0.11&52.9&46.8\\
			GNN-RL\cite{yu2022topology}&93.49&54.0&—\\
			SOSP\cite{nonnenmacher2021sosp}&93.27±0.51&57.0&61.0\\
			FP-AGL\cite{kim2022fp}&93.49&60.9&—\\
			\textbf{Ours}($C$=0.385)&\textbf{93.52±0.18}&\textbf{61.1}&52.5\\
			\hdashline
			FTWT\cite{elkerdawy2022fire}&92.63&66.0&—\\
			CHIP\cite{sui2021chip}&92.05&72.3&71.8\\
			Hrank\cite{lin2020hrank}&90.72&74.1&68.1\\
			HPSE\cite{zhang2022hierarchical}&91.51&74.2&—\\
			\textbf{Ours}($C$=0.25)&\textbf{92.84±0.21}&\textbf{74.8}&66.8\\
			\hline\hline
			ResNet-110&93.50&0.0&0.0\\
			PFEC\cite{li2016pruning}&93.30&38.6&32.4\\
			PFS\cite{wang2020pruning}&93.69±0.28&40.0&—\\
			SFP\cite{he2018soft}&93.38±0.30&40.8&—\\
			Hrank\cite{lin2020hrank}&94.23&41.2&39.4\\
			FPGM\cite{he2019filter}&93.74±0.10&52.3&—\\
			\textbf{Ours}($C$=0.47)&\textbf{94.03±0.26}&\textbf{52.5}&42.7\\
			\hdashline
			LFPC\cite{he2020learning}&93.79±0.38&60.3&—\\
			HPSE\cite{zhang2022hierarchical}&93.79&60.6&—\\
			DECORE\cite{alwani2022decore}&93.50&61.8&64.8\\
			Hrank\cite{lin2020hrank}&92.65&68.6&69.2\\
			\textbf{Ours}($C$=0.31)&\textbf{93.90±0.14}&\textbf{68.7}&66.2\\
			\hdashline
			HPSE\cite{zhang2022hierarchical}&93.26&73.4&—\\
			DECORE\cite{alwani2022decore}&92.71&76.9&79.6\\
			\textbf{Ours}($C$=0.22)&\textbf{93.18±0.17}&\textbf{77.8}&70.6\\
			\hline
		\end{tabular}
	\end{center}
\end{table}

{\bf{VGG-16.}} The compression results of VGG-16 are displayed in Table \ref{vgg16}. Compared with post-training pruning methods HRank and DECORE, our method without pre-training is better in various acceleration levels. Compared with CC and Hinge which combine pruning and low-rank decomposition, our method also obtains better performance (65.7\% vs. 60.7\% vs. 39.1\% in FLOPs reduction, and 94.36\% vs. 94.09\% vs. 94.02\% in top-1 accuracy). Compared with dynamic pruning method FTWT, with almost the same accuracy, our method obtains significantly more FLOPs reduction (73.8\% vs. 56.0\%). Compared with PFEC which also uses the ${\ell}_1$-norm metric, where only 93.40\% top-1 accuracy is obtained, our method obtains a better result of 94.32\% with significantly more FLOPs reduction, which demonstrates the superiority of introducing the bypasses and using the proposed non-uniform pruning strategy. Therefore, our method demonstrates its ability to accelerate a neural network with a plain structure.

{\bf{ResNet-56/110/20/32.}} The compression results of ResNet-56/110/20/32 are displayed in Tables \ref{resnet56/110} and \ref{resnet20/32}. We start with ResNet-56. Compared with the pruning at initialization method PFS, our method further reduces FLOPs by 2.5\%, while maintaining higher accuracy (93.72(±0.16)\% vs. 93.05(±0.19)\%). Besides, compared with the pruning during training method DPFPS, our method further improves accuracy by 0.32\%, while maintaining more FLOPs reduction (61.1\% vs. 52.9\%). Compared with NPPM which also uses FLOPs constraints, with more FLOPs reduction, our method without pre-training obtains significantly better accuracy (93.72(±0.16)\% vs. 93.40\%), which demonstrates the superiority of our pruning strategy combining learnable thresholds and FLOPs constraints. Compared with LFPC which learns filter pruning criteria, with almost the same accuracy, our method further reduces FLOPs by 5.4\%. 

On ResNet-110 with more redundancy, our method leads to significant improvement in accuracy over the baseline model (94.03(±0.26)\% vs. 93.50\%) with around 52.5\% FLOPs and 42.7\% parameters reduction. Compared with SFP and FPGM with the same pruning rate for all layers, our method achieves higher accuracy (93.90(±0.14)\% vs. 93.38(±0.30)\% vs. 93.74(±0.10)\%) and saves more computing resources (68.7\% vs. 40.8\% vs. 52.3\%), which show that our method is more effective. Compared with HPSE with a complex pruning criterion based on the characteristics of feature maps, our method using the simple ${\ell}_1$-norm metric is still comparable in various acceleration levels, which demonstrates the superiority of our method.

\begin{table}
\begin{center}
\caption{Compression results of ResNet-20/32 on CIFAR-10.\label{resnet20/32}}
\begin{tabular}{m{1.6cm}|c|c|c}
\hline
 Method & Top-1(\%) &   FLOPs ↓(\%)  & Params ↓(\%)  \\
\hline\hline
ResNet-20&92.20&0.0&0.0\\
SFP\cite{he2018soft}&90.83±0.31&42.2&—\\
FPGM\cite{he2019filter}&91.09±0.10&42.2&—\\
PFS\cite{wang2020pruning}&90.55±0.14&50.0&—\\
GNN-RL\cite{yu2022topology}&91.31&51.0&—\\
Hinge\cite{li2020group}&91.84&54.5&55.5\\
\textbf{Ours}($C$=0.4)&\textbf{92.22±0.19}&\textbf{59.7}&52.8\\
\hdashline
FP-AGL\cite{kim2022fp}&90.11&60.7&—\\
\textbf{Ours}($C$=0.28)&\textbf{91.24±0.24}&\textbf{71.8}&58.9\\
\hline\hline
ResNet-32&92.63&0.0&0.0\\
SFP\cite{he2018soft}&92.08±0.08&41.5&—\\
GNN-RL\cite{yu2022topology}&92.58&51.0&—\\
LFPC\cite{he2020learning}&92.12±0.08&52.6&—\\
FPGM\cite{he2019filter}&91.93±0.03&53.2&—\\
\textbf{Ours}($C$=0.4)&\textbf{93.21±0.19}&\textbf{59.6}&49.1\\
\hdashline
FP-AGL\cite{kim2022fp}&91.86&60.8&—\\
\textbf{Ours}($C$=0.28)&\textbf{92.32±0.21}&\textbf{71.8}&60.8\\
\hline
\end{tabular}
\end{center}
\end{table}

Table \ref{resnet20/32} shows the performance of the different methods of compressing ResNet-20/32. Compressing the relatively lightweight ResNet20/32 is more challenging than compressing the ResNet-56/110. On ResNet-20, compared with GNN-RL which finds suitable hidden layers’ pruning rates using reinforcement learning, our method further improves accuracy by 0.91\% and reduces FLOPs by 8.7\%. On ResNet-32, with around 60.0\% FLOPs reduction, our method leads to significant improvement in accuracy over the baseline model (93.21(±0.19)\% vs. 92.63\%), which is significantly different from other advanced compression algorithms, highlighting the effectiveness of our method. Therefore, our method demonstrates that it is especially suitable for compressing neural networks with residual blocks.

\begin{table}
\begin{center}
\caption{Compression results of DenseNet-40 on CIFAR-10.\label{densenet40}}
\begin{tabular}{m{1.8cm}|c|c|c}
\hline
 Method & Top-1(\%) &   FLOPs ↓(\%)  & Params ↓(\%)  \\
\hline\hline
Baseline&94.81&0.0&0.0\\
DECORE\cite{alwani2022decore}&94.59&39.4&46.0\\
Hrank\cite{lin2020hrank}&94.24&40.8&36.5\\
Hinge\cite{li2020group}&94.67&44.4&27.5\\
CC\cite{li2021towards}&\textbf{94.67}&47.0&51.9\\
HPSE\cite{zhang2022hierarchical}&94.38&48.0&—\\
\textbf{Ours}($C$=0.5,in)&94.56±0.17&49.5&46.6\\
\textbf{Ours}($C$=0.5,out)&94.51±0.29&\textbf{49.6}&48.2\\
\hdashline
DECORE\cite{alwani2022decore}&94.04&54.7&65.0\\
NS\cite{liu2017learning}&94.35&55.0&65.2\\
CC\cite{li2021towards}&\textbf{94.40}&60.4&64.4\\
Hrank\cite{lin2020hrank}&93.68&61.0&53.8\\
HPSE\cite{zhang2022hierarchical}&93.88&61.5&—\\
\textbf{Ours}($C$=0.38,in)&94.24±0.25&\textbf{61.6}&62.4\\
\hline
\end{tabular}
\end{center}
\end{table}

{\bf{DenseNet-40.}} Table \ref{densenet40} shows the performance of the different methods of compressing DenseNet-40, where `in' denotes channel pruning and `out' denotes filter pruning. Compared with the state-of-the-art pruning methods HRank, DECORE and HPSE, our method is better than them in various acceleration levels. Compared with CC and Hinge requiring pre-trained models, with slightly more FLOPs reduction, although our average accuracy is worse than the results they report, our best result is better than theirs (94.76\% vs. 94.67\% vs. 94.67\%). Overall, our method has the potential to better accelerate networks with dense blocks.

\begin{table}
\begin{center}
\caption{Compression results of GoogleNet on CIFAR-10.\label{googlenet}}
\begin{tabular}{m{1.7cm}|c|c|c}
\hline
 Method & Top-1(\%) &   FLOPs ↓(\%)  & Params ↓(\%)  \\
\hline\hline
Baseline&95.05&0.0&0.0\\
DECORE\cite{alwani2022decore}&95.20&19.8&23.0\\
PFEC\cite{li2016pruning}&94.54&32.9&42.9\\
CC\cite{li2021towards}&95.18&50.0&54.0\\
\textbf{Ours}($C$=0.49)&\textbf{95.57±0.08}&\textbf{50.5}&52.6\\
\hdashline
Hrank\cite{lin2020hrank}&94.53&54.9&55.4\\
CC\cite{li2021towards}&94.88&59.9&63.3\\
\textbf{Ours}($C$=0.39)&\textbf{95.40±0.12}&\textbf{60.6}&61.1\\
\hdashline
Hrank\cite{lin2020hrank}&94.07&70.4&69.8\\
DECORE\cite{alwani2022decore}&94.51&78.5&80.9\\
\textbf{Ours}($C$=0.21)&\textbf{94.99±0.08}&\textbf{78.8}&79.1\\
\hline
\end{tabular}
\end{center}
\end{table}

{\bf{GoogleNet.}} In Table \ref{googlenet}, we analyze the performance of the different methods on GoogleNet. Because GoogleNet has large amount of redundancy, compressing it is especially easy. 1$\times$1 convolutions in GoogleNet possess only sparse paths for our method. With almost the same accuracy as the baseline model, our method obtains around 79\% FLOPs and parameters reduction. Besides, compared with HRank and CC, our method uses much fewer FLOPs (39.4\% vs. 45.1\% vs.40.1\%), but achieves higher accuracy (95.40(±0.12)\% vs. 94.53\% vs. 94.88\%), which is very encouraging. Therefore, it demonstrates that our method can be effectively applied to compressing neural networks with inception modules.

\subsection{Experimental Results on ImageNet}

\begin{table}
\begin{center}
\caption{Compression results of ResNet-18/34 on ImageNet.\label{resnet18/34}}
\begin{tabular}{m{1.6cm}|c|c|c}
\hline
Method & Top-1(\%) &   Top-5(\%)  & FLOPs ↓(\%)  \\
\hline\hline
ResNet-18&69.76&89.08&0.0\\
SOSP\cite{nonnenmacher2021sosp}&68.78±0.98&—&29.0\\
SFP\cite{he2018soft}&67.10&87.78&41.8\\
FPGM\cite{he2019filter}&68.34&88.53&41.8\\
\textbf{Ours}($C$=0.56)&\textbf{70.43±0.07}&\textbf{89.77±0.03}&\textbf{43.5}\\
\hdashline
DCP\cite{zhuang2018discrimination}&67.36&87.63&46.2\\
GNN-RL\cite{yu2022topology}&68.66&—&51.0\\
FTWT\cite{elkerdawy2022fire}&67.49&—&51.6\\
\textbf{Ours}($C$=0.44)&\textbf{69.97±0.04}&\textbf{89.48±0.14}&\textbf{55.6}\\
\hline\hline
ResNet-34&73.31&91.42&0.0\\
SFP\cite{he2018soft}&71.83&90.33&41.1\\
FPGM\cite{he2019filter}&72.54&91.13&41.1\\
FP-AGL\cite{kim2022fp}&72.72&—&43.1\\
DPFPS\cite{ruan2021dpfps}&72.25&90.80&43.3\\
\textbf{Ours}($C$=0.56)&\textbf{72.91±0.13}&\textbf{91.18±0.02}&\textbf{43.5}\\
\hdashline
ResNet-18\cite{he2016deep}&69.76&89.08&50.0\\
FTWT\cite{elkerdawy2022fire}&71.71&—&52.2\\
\textbf{Ours}($C$=0.47)&\textbf{72.55±0.16}&\textbf{91.05±0.11}&\textbf{52.5}\\
\hline
\end{tabular}
\end{center}
\end{table}

{\bf{ResNet-18/34.}} We also conduct experiments for ResNet-18 and ResNet-34 on the challenging ImageNet dataset, as shown in Table \ref{resnet18/34}. We can see the similar conclusion in CIFAR-10 dataset, i.e., our method is almost superior to compared methods in all aspects. Specifically, Compared with SFP (67.10\% top-1) and FPGM (68.34\% top-1), our method achieves higher accuracy (70.43\% top-1) with more FLOPs reduction (43.5\%) on ResNet-18. Compared with DCP requiring pre-trained models, our method provides significantly higher accuracy with more FLOPs reduction (69.97(±0.04)\% vs. 67.36\% in top-1 accuracy, and 55.6\% vs. 46.2\% in FLOPs reduction) on ResNet-18. Compared with GNN-RL requiring total 240 epochs, with 120 epochs, our method obtains better performance on ResNet-18, which show that our method is more effective. Compared with FP-AGL and DPFPS, with higher accuracy, our method obtains similar FLOPs reduction on ResNet-34. Compared with ResNet-18, with more FLOPs reduction, our method leads to significantly higher accuracy on ResNet-34, which demonstrates that our method is able to generate a good sub-network. FTWT is a dynamic pruning method and expected to outperform static pruning methods, but our method still outperforms it on ResNet-18 and ResNet-34. Hence, our method also works well on complex datasets.

\begin{table}
	\begin{center}
		\caption{Ablation experimental results of the bypass.\label{bypass}}
		\begin{tabular}{m{1.1cm}|c|c|c|c}
			\hline
			Model&$C$&Method & Top-1(\%) &   FLOPs ↓(\%) \\
			\hline\hline
			\multirow{4}*{VGG-16} &\multirow{4}*{0.34}&LAPP-S&93.70±0.24&65.7\\
			&&Model-V2&94.20±0.15&65.5\\
			&&LAPP-V1&93.65±0.13&65.7\\
			&&\textbf{LAPP}&\textbf{94.36±0.08}&65.7\\
			\hline\hline
			\multirow{4}*{ResNet-56} &\multirow{4}*{0.33}&LAPP-S&92.93±0.21&66.7\\
			&&Model-V2&92.84±0.11&66.3\\
			&&LAPP-V1&93.09±0.17&66.7\\
			&&\textbf{LAPP}&\textbf{93.49±0.07}&66.7\\
			\hline\hline
			\multirow{4}*{ResNet-20} &\multirow{4}*{0.4}&LAPP-S&90.70±0.13&59.7\\
			&&Model-V2&91.68±0.17&59.3\\
			&&LAPP-V1&91.71±0.48&59.7\\
			&&\textbf{LAPP}&\textbf{92.22±0.19}&59.7\\
			\hline
		\end{tabular}
	\end{center}
\end{table}

\subsection{Ablation Study}

{\bf{Effect of the Bypass.}} As mentioned earlier, the bypasses compensate for the sparse paths. We conduct ablation experiments on CIFAR-10 with VGG-16, ResNet-56 and ResNet-20 to show the compensation effects of the bypasses, as shown in Table \ref{bypass}. Here, the blocks inspired by MobileNetV1 \cite{howard2017mobilenets} and MobileNetV2 \cite{sandler2018mobilenetv2} are named V1 and V2, respectively. The block V1 sequentially contains two layers of small convolutions: depthwise convolution and $1\times1$ convolution. `LAPP-S' denotes only using sparse paths in SBC modules, `Model-V2' denotes only using the bypasses containing the block V2, and `LAPP-V1' and 'LAPP' denote using the two paths and respectively take the block V1 and V2 as the bypass. We adjust the hyperparameter $d^l$ of each bypass for Model-V2 to obtain the same FLOPs reduction as other methods. Besides, we use the exact same training configurations for all methods. Unlike LAPP-V1 and LAPP, for LAPP-S without the bypasses, when output channels of the layer are pruned, pruning the corresponding input channels of the next layer needs to be taken into account. 

From Table \ref{bypass}, LAPP-V1 and LAPP with the bypasses outperform LAPP-S in almost all the cases, which demonstrates the compensation effects of the bypasses. What's more, the more compact the network, and the more significant the compensation effect of the bypass, especially for networks with residual blocks. For example, when compressing more challenging ResNet-20, LAPP surpasses LAPP-S by 1.52\% in top-1 accuracy with the similar FLOPs reduction. Besides, for compressing VGG-16, ResNet56 and ResNet-20, LAPP surpasses Model-V2 by 0.16\%, 0.65\%, and 0.54\% in top-1 accuracy, respectively. This proves that the bypasses and the sparse paths can compensate for each other, which results in better compression performance. Last but not least, LAPP outperforms LAPP-V1 in all the cases. This proves that the block V2 with stronger expressive power has better compensation effect than the block V1.

\begin{table}
	\begin{center}
		\caption{Ablation experimental results of the Pruning Strategy.`UP' denotes uniform pruning.\label{PS}}
		\begin{tabular}{m{1.1cm}|c|c|c|c}
			\hline
			Model&$C$&Method & Top-1(\%) &   FLOPs ↓(\%)  \\
			\hline\hline
			\multirow{4}*{VGG-16}&\multirow{2}*{0.34}&LAPP-UP&94.20±0.11&65.8\\
			&&\textbf{LAPP}&\textbf{94.36±0.08}&65.7\\
			\cline{2-5}
			&\multirow{2}*{0.18}&LAPP-UP&93.31±0.12&81.8\\
			&&\textbf{LAPP}&\textbf{93.54±0.23}&81.9\\
			\hline\hline
			\multirow{4}*{ResNet-56}&\multirow{2}*{0.47}&LAPP-UP&93.34±0.18&50.1\\
			&&\textbf{LAPP}&\textbf{93.72±0.16}&\textbf{52.5}\\
			\cline{2-5}
			&\multirow{2}*{0.33}&LAPP-UP&92.84±0.13&65.2\\
			&&\textbf{LAPP}&\textbf{93.49±0.07}&\textbf{66.7}\\
			\hline\hline
			\multirow{2}*{ResNet-20}&\multirow{2}*{0.4}&LAPP-UP&91.25±0.18&57.7\\
			&&\textbf{LAPP}&\textbf{92.22±0.19}&\textbf{59.7}\\
			\hline
		\end{tabular}
	\end{center}
\end{table}

{\bf{Effect of the Pruning Strategy.}} Here we conduct ablation experiments on CIFAR-10 with VGG-16, ResNet-56 and ResNet-20 to show the effects of the pruning strategy under different target FLOPs compression rates, as shown in Table \ref{PS}. Using the same filter importance metric, we compare our pruning strategy with the uniform pruning strategy at initialization. We use `UP' to denote uniform pruning before training starts. We manually set the almost same pruning rates for all sparse paths to reach target FLOPs compression rates. Other training configurations are the exact same as LAPP. LAPP with our pruning strategy outperforms LAPP-UP using uniform pruning in all the cases. For example, when compressing ResNet-56, LAPP surpasses LAPP-UP by 0.38\% and 0.65\% in top-1 accuracy respectively under different target FLOPs compression rates 47\% and 33\%. The lower the target FLOPs compression rate, the more effective our pruning strategy will be. This proves that our pruning strategy is able to generate a relatively better sub-network.

\section{Conclusion}
\noindent In this paper, we integrate learnable thresholds and FLOPs constraints into an effective and efficient pruning strategy. The pruning strategy can gradually prune the network and automatically determine the appropriate pruning rates for each layer during initial training of a few epochs from scratch. Besides, we introduce an additional lightweight bypass for each convolutional layer to compensate the lost information and capacity of the pruned layer during training. The effectiveness of the proposed method is evaluated on the benchmark datasets CIFAR-10 and ImageNet using the mainstream CNN models, and the results show its superior performance compared with other advanced compression algorithms. 

{\small
\bibliographystyle{IEEETran}
\bibliography{references_ieee}
}

\end{document}